\pdfoutput=1

\documentclass[11pt]{article}

\usepackage{acl}

\usepackage{times}
\usepackage{latexsym}

\usepackage[T1]{fontenc}

\usepackage[utf8]{inputenc}

\usepackage{microtype}

\usepackage{inconsolata}

\usepackage{graphicx}

\usepackage{booktabs}
\usepackage{pifont}
\usepackage{enumitem}
\usepackage{hyperref}
\usepackage{graphicx}
\usepackage{adjustbox}
\usepackage{rotating}
\usepackage{multirow}
\usepackage{amssymb}
\usepackage{amsmath}

\usepackage{cleveref} 
\usepackage{tikz}
\usetikzlibrary{positioning, arrows.meta}
\usepackage{comment}
\usepackage{float}
\usepackage{lipsum}  
\usepackage[yyyymmdd,hhmmss]{datetime}
\usepackage{array}
\newcolumntype{L}[1]{>{\raggedright\let\newline\\\arraybackslash\hspace{0pt}}m{#1}}
\newcolumntype{C}[1]{>{\centering\let\newline\\\arraybackslash\hspace{0pt}}m{#1}}
\newcolumntype{R}[1]{>{\raggedleft\let\newline\\\arraybackslash\hspace{0pt}}m{#1}}
\usepackage{siunitx} 
\usepackage{subcaption}

\usepackage{comment}
\usepackage{multirow}
\usepackage{siunitx}
\usepackage{amssymb}
\usepackage{pifont}
\newcommand{\xmark}{\ding{55}}%

\usepackage{etoolbox}
\usepackage{pgf} 
\usepackage{xcolor} 
\usepackage{colortbl}

\newcommand{\gradientcell}[6]{
    \ifdimcomp{#1pt}{>}{#3 pt}{\cellcolor{#5!100.0!#4!#6}#1}{
    \ifdimcomp{#1pt}{<}{#2 pt}{\cellcolor{#5!0.0!#4!#6}#1}{
         \pgfmathparse{int(round(100*(#1/(#3-#2))-(#2 *(100/(#3-#2)))))}
        \xdef\tempa{\pgfmathresult}
        \cellcolor{#5!\tempa!#4!#6}#1
    }}
 }

\newcommand{\oAc}[1]{\gradientcell{#1}{.56}{.86}{cyan}{yellow}{70}}
\newcommand{\oR}[1]{\gradientcell{#1}{.53}{.69}{cyan}{yellow}{70}}

\newcommand{\cAc}[1]{\gradientcell{#1}{.48}{.84}{cyan}{yellow}{70}}
\newcommand{\cR}[1]{\gradientcell{#1}{.53}{.64}{cyan}{yellow}{70}}

\newcommand{\lAc}[1]{\gradientcell{#1}{.58}{.91}{cyan}{yellow}{70}}
\newcommand{\lR}[1]{\gradientcell{#1}{.53}{.65}{cyan}{yellow}{70}}

\newcommand{\sAc}[1]{\gradientcell{#1}{.69}{.90}{cyan}{yellow}{70}}
\newcommand{\sR}[1]{\gradientcell{#1}{.53}{.65}{cyan}{yellow}{70}}

\newcommand{\oC}[1]{\gradientcell{#1}{.78}{.90}{cyan}{yellow}{70}}
\newcommand{\cC}[1]{\gradientcell{#1}{.62}{.92}{cyan}{yellow}{70}}
\newcommand{\sC}[1]{\gradientcell{#1}{.26}{.60}{cyan}{yellow}{70}}

\newcommand{\oCC}[1]{\gradientcell{#1}{.24}{.90}{cyan}{yellow}{70}}
\newcommand{\cCC}[1]{\gradientcell{#1}{.42}{.74}{cyan}{yellow}{70}}
\newcommand{\lCC}[1]{\gradientcell{#1}{.5}{.89}{cyan}{yellow}{70}}
\newcommand{\sCC}[1]{\gradientcell{#1}{.61}{.92}{cyan}{yellow}{70}}

\title{Enhancing Conceptual Understanding in Multimodal Contrastive Learning through Hard Negative Samples}

\author{Philipp J. Rösch$^1$, Norbert Oswald$^1$, Michaela Geierhos$^1$, \and Jindřich Libovický$^2$ \\
$^1$Bundeswehr University Munich, Germany \\
$^2$Faculty of Mathematics and Physics, Charles University, Czech Republic \\
\texttt{\{philipp.roesch,norbert.oswald,michaela.geierhos\}@unibw.de} \\ \texttt{libovicky@ufal.mff.cuni.cz}
}

\begin{document}
\maketitle
\begin{abstract}
Current vision-language models leveraging contrastive learning often face limitations in developing fine-grained conceptual understanding.
This is due to random negative samples during pretraining, causing almost exclusively very dissimilar concepts to be compared in the loss function. Consequently, the models struggle with fine-grained semantic differences.
To address this problem, we introduce a novel pretraining method incorporating synthetic hard negative text examples. The hard negatives replace terms corresponding to visual concepts, leading to a more fine-grained visual and textual concept alignment.
Further, we introduce InpaintCOCO, a new challenging dataset for assessing the fine-grained alignment of colors, objects, and sizes in vision-language models. We created the dataset using generative inpainting from COCO images by changing the visual concepts so that the images no longer match their original captions.
Our results show significant improvements in fine-grained concept understanding across various vision-language datasets, including our InpaintCOCO dataset.
\end{abstract}

\begin{figure*}[!ht]
\small
\centering
\begin{tabular}{p{.1cm}p{2.cm}|p{2.7cm}|p{2.7cm}|p{2.7cm}|p{2.7cm} }
$I_{1}$ & \multirow{2}{*}{\includegraphics[width=1.8cm]{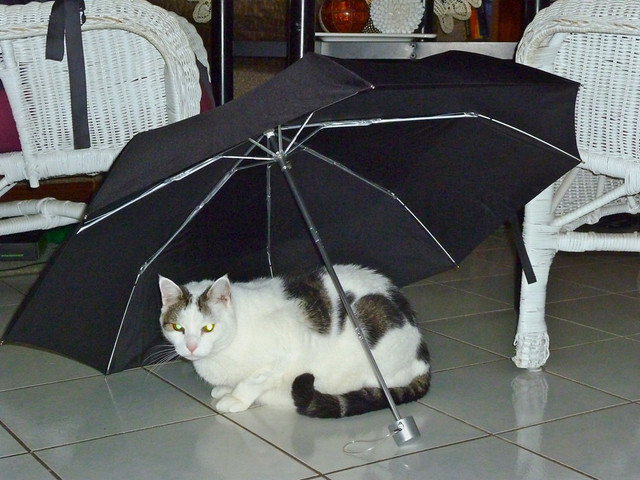}} & $T^{pos}_{1}$ & $T^{neg}_{2}$ & $T^{neg}_{3}$ & $T^{hn}_{1}$ \\ 
& & a white \underline{cat} sits under a black open \underline{umbrella}. & a \underline{person} standing on a loading \underline{platform} next to a \underline{train}. & a \underline{man} leans near the \underline{door} of a creatively painted \underline{tour bus}. & a \textbf{brown} \underline{cat} sits under a black open \underline{umbrella}.  \\
\end{tabular}
\caption{Classical contrastive learning approaches use $(I_{1}, T^{pos}_{1})$ as positive pairs in combination with negative samples like $T^{neg}_{2}$ and $T^{neg}_{3}$ to learn an image-text alignment.
A bag of words (e.g., nouns) is often sufficient to extract the correct text that matches a given image, resulting in only broad concepts learned.
We also use hard negatives like $T^{hn}_{1}$ so that fine-grained semantic concepts are learned for visual and textual alignment.}
\label{fig:examples}
\end{figure*}

\section{Introduction}

Recent advancements in vision-language (VL) modeling have demonstrated the effectiveness of contrastive learning in various multimodal tasks \citep{radford2021learning, jia2021scaling, yao2021filip}. 
However, this training method does not provide sufficient training signals for several important visual concepts \cite{zhao2023vlchecklist}.
We attributed it to the objective function's use of random and, therefore, too dissimilar negative samples, which prevents the model from learning fine-grained semantic representations of the concepts.

Therefore, we propose a novel approach to address the issue of poorly represented concepts in contrastive learning.
We introduce a mechanism to incorporate hard negative samples into the contrastive learning loss. Specifically, we generate synthetic hard negative samples by substituting keywords in the captions of original image-text pairs, disrupting the alignment between the image content and its description.

This paper presents three key contributions:

\textbf{(i)} We present a novel method for using hard negative samples in the contrastive learning objective, allowing the model to focus on refining its understanding of concepts.

\textbf{(ii)} By introducing hard negative samples in the language component, we compel the model to learn proper visual and language alignment.  Our approach improves multimodal performance, although it operates exclusively on the language side of the model.

\textbf{(iii)} To evaluate the model from the visual perspective, we create a challenge set with over 1,260 adversarial examples by using generative image inpainting. 
This dataset serves as a comprehensive benchmark, allowing us to assess the model's ability to validate its conceptual understanding.
This is because the image was created in a standardized setting in which only a small part was changed.

In this work, we conduct extensive evaluations of four basic concepts -- color, object, location, and size. These concepts were selected as examples to demonstrate the effectiveness and robustness of our proposed approach in capturing nuanced semantic relations, but it is important to note that the choice of concepts is flexible and can be tailored to specific applications. 
Furthermore, our methodology is easy to construct, requiring only minimal domain expertise and the simple usage of regular expressions. 
This study shows that simple tweaks in contrastive learning can significantly enhance multimodal understanding and model performance.

\begin{figure*}[h!]
\centering
\includegraphics[clip, trim=5.1cm 0cm 0.1cm 0cm, width=.96\textwidth]{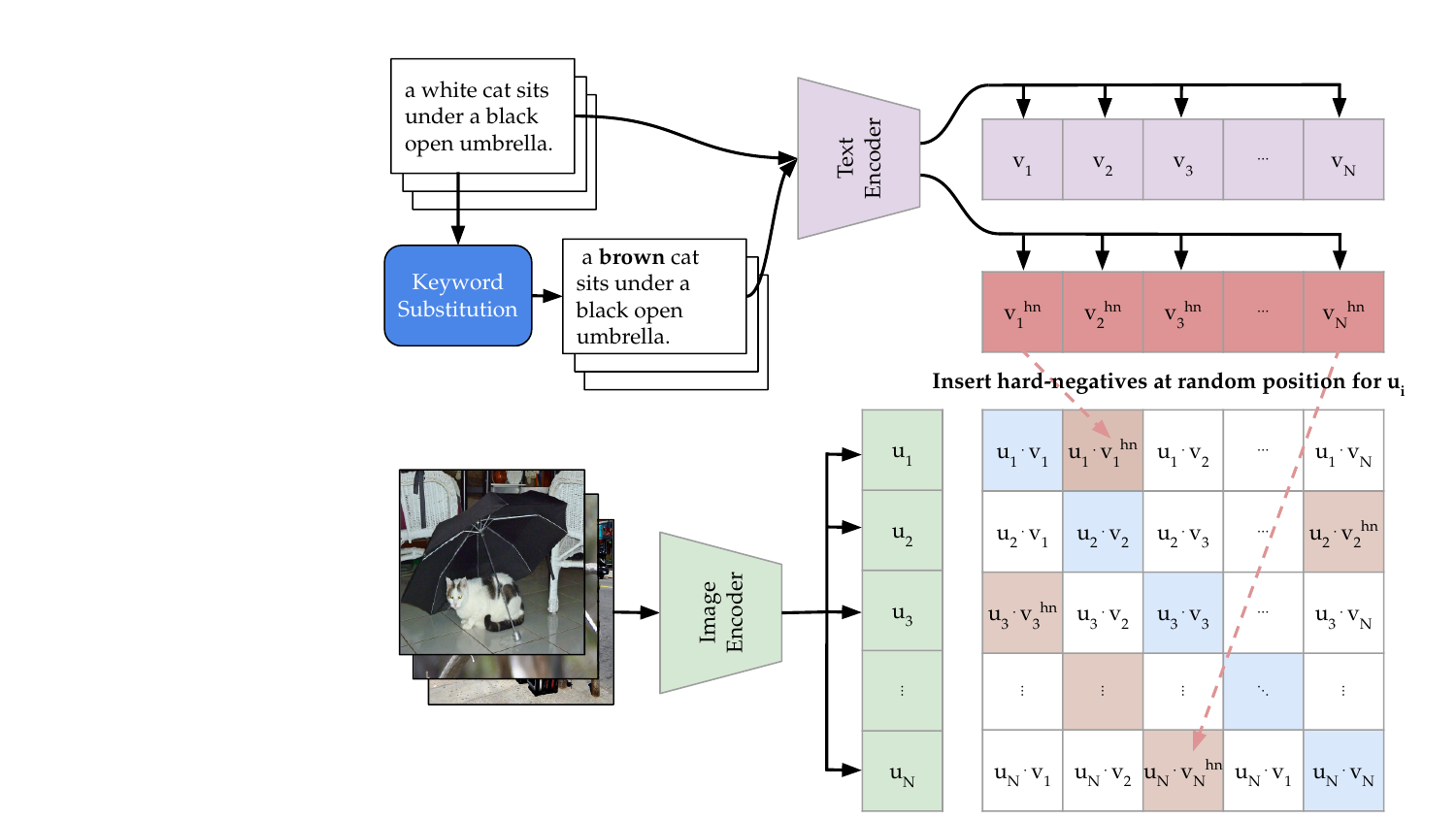} 
\caption{Hard negative contrastive learning: Keyword substitution produces hard negative text samples, which are then randomly injected for each image $u_i$, replacing a simple negative sample in InfoNCE loss.}
\label{fig:hnclip}
\end{figure*}

\section{Vision-Language Representation Learning}

\paragraph{Contrastive Learning.}
The objective of contrastive representation learning is to learn representations that are close to each other for similar samples and distant from each other for dissimilar samples. 
While many objectives originally addressed a single modality \citep{Chopra2005CL, Schroff_2015, sohn2016improved, oord2018representation}, the idea can also be extended to multimodal training as well.

A successful example of multimodal learning is CLIP \citep{radford2021learning}. 
CLIP is a Transformer-based model that consists of an image encoder and a text encoder, which are trained simultaneously.
The objective is to maximize the cosine similarity of the image and text embeddings from the correct image-text pairs and to minimize the similarity between the incorrect pairs. 
A batch of $N$ training samples (i.e., matching image-text pairs) results in a similarity matrix for each image-text combination. 
The main diagonal indicates the correct pair matches; the remaining entries correspond to negative entries.
The symmetric cross-entropy loss is applied on $N \times N$ similarity scores. 

This heuristical construction of negative samples has several issues.
Negative captions can still match the given image in some cases, especially if the text is short and lacks details.
Additionally, the negative pairs are often very dissimilar, which causes the model to decide only on coarse-grained features. 

We illustrate this in \Cref{fig:examples} with text-image pair $I_1$ and $T_1^{pos}$.
The original learning process only uses negative text samples, such as  $T^{neg}_{2}$ and $T^{neg}_{3}$. 
Here, it is sufficient that the model can assign or negate objects from texts (i.e., nouns)  to the objects in the image. 
The fact that no ``person'' and no ``door'' are present in image $I_1$ is sufficient for the model to discard these image descriptions. 
As a result, fine-grained concepts like object-color alignment, size, or spatial details (``cat under umbrella'') are not necessary for reaching low loss.
In this example, the model can rely solely on the presence and absence of specific objects. In this scenario, the caption can be seen as a bag of words without linguistic structure.   

We address this problem by creating new hard negative data samples to learn more fine-grained concepts. 
See \S~\ref{sec:contrastive-learning-with-hn} for details.

\paragraph{Related work.}
Several approaches incorporate hard negative samples in multimodal learning. 

\citet{radenovic2023filtering} use importance sampling (based on \citealp{robinson2021contrastive}), which up-samples hard negative samples and down-samples or ignores simple negatives. 
Yet, they reweight the simple negative samples per batch only but do not create new, challenging samples requiring fine-grained understanding.

\citet{rosch_libovicky2022probing} propose keyword permutation to create hard negative samples to learn spatial concepts. 
They added a spatial understanding classifier as an auxiliary pretraining objective and evaluated the models on visual question answering (VQA).  
Their model is based on LXMERT \citep{tanbansal2019lxmert} and not on a contrastive learning approach like CLIP.

\citet{doveh2023teaching} generate hard negatives using a rule-based procedure where they replace keywords. 
Moreover, they implement an approach where they randomly parse parts of speech and fill the mask using a BERT encoder with a plausible but wrong word. 
Unlike us, they do not incorporate hard negatives into the similarity matrix but use an auxiliary loss summed with the original contrastive loss. 
They use four distinct loss functions in total, introducing an additional layer of complexity to the overall procedure.
In contrast, our approach uses the original loss function, with minimal modifications limited to the text inputs.

\section{Contrastive Learning with Hard Negatives}
\label{sec:contrastive-learning-with-hn}


We present a novel contrastive learning approach using sampled negative pairs and artificially generated textual hard negatives.

Instead of training with one positive sample and several weak negative samples, we create a scenario where models also minimize the similarity to hard negative \emph{textual} samples.
This forces the model to learn fine-grained concepts during training.

We use keyword substitutions for different concepts to break the correct meaning of an image caption. See \Cref{fig:examples} for an example of the concept \textit{color}. 
As a result, the new caption still lists, e.g., correct objects (e.g., ``cat'' and ``umbrella'') and actions (e.g., ``cat sits'') from the image but is no longer correct. 
We call this a ``hard negative sample''. 
Using these samples during training, we ensure that fine-grained concepts are learned.  
The idea to inject hard negative samples in contrastive learning is highlighted in \Cref{fig:hnclip}.

\subsection{Creating Hard Negative Text Samples}
\label{sec:CreateHN}

\begin{figure*}
\centering
\footnotesize
\begin{tikzpicture}[>=latex, node distance=2cm]

    \node (box1) [draw, rectangle, minimum width=2.75cm, minimum height=3cm, align=center] at (0,0) {%
        COCO sample\\
        \includegraphics[width=2cm,height=2cm]{assets/fig3/image.png}\\
        A \textbf{red} fire hydrant  \\  is on the sidewalk \\  next to a sign.
    };

    \draw[->] (box1.east) -- node[below,align=center,rotate=90,anchor=west] {Mask\\``fire hydrant''} ++(1.3,0);

    \node (box2) [draw, rectangle, minimum width=2.5cm, minimum height=3cm, right of=box1, node distance=4cm, align=center] {%
        Mask\\
        \includegraphics[width=2cm,height=2cm]{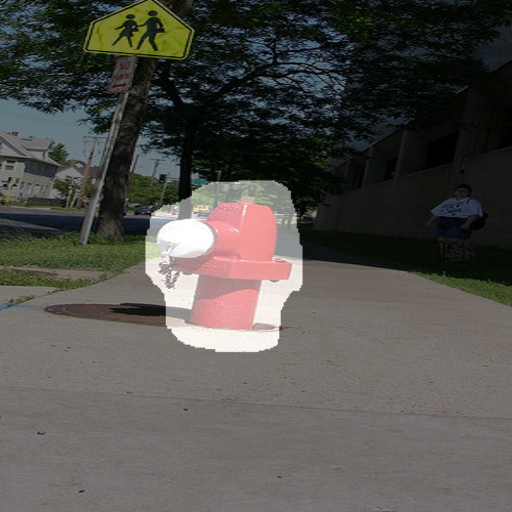}
    };

    \draw[->] (box2.east) -- node[below,align=center,rotate=90,anchor=west]{Inpaint ``yellow \\ fire hydrant''} ++(1.45,0);

    \node (box3) [draw, rectangle, minimum width=2.5cm, minimum height=3cm, right of=box2, node distance=4cm, align=center] {%
        Inpainted\\
        \includegraphics[width=2cm,height=2cm]{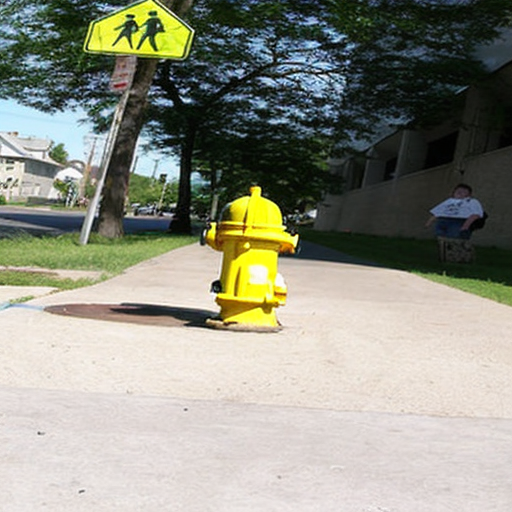}
    };

    \draw[->] (box3.east) -- node[below,align=center,rotate=90,anchor=west] {Add Caption} ++(1.25,0);
    
    \node (box4) [draw, rectangle, minimum width=3cm, minimum height=3cm, right of=box3, node distance=4cm, align=center] {%
            New sample\\
            \includegraphics[width=2cm,height=2cm]{assets/fig3/inpainted_image.png}\\
            A \textbf{yellow} fire hydrant  \\  is on the sidewalk \\  next to a sign.
    };

\end{tikzpicture}
\caption{Create hard negative image samples using open vocabulary segmentation for the masking prompt and text-to-image generation for the inpainting prompt. Additionally, a new correct caption is created manually. The InpaintCOCO dataset was created for concepts \textit{object}, \textit{color}, and \textit{size}.}
\label{fig:inpainting}
\end{figure*}

For various concepts, we replace specific keywords using a \textit{regex}-based tool. 
For example, we replace ``white cat'' with ``brown cat'' or ``cat'' with ``dog''.
We create substitution heuristics for four different concepts, namely \textit{colors}, \textit{objects}, \textit{size}, \textit{location}:
\begin{itemize}[noitemsep,topsep=0pt] 
    \item For \textit{color}, any of the 9 most occurring color names in COCO can be replaced by any other color name.
    \item For \textit{objects}, any of 80 object names can be replaced by any other. The 80 words originate from COCO object categories.
    \item For \textit{location} keywords we use 12 one-to-one substitution relations. 
    \item For \textit{size} keywords we use 11 one-to-one substitution relations.
\end{itemize}
The full list of heuristics is shown in \Cref{tab:heuristic-permutation} in the Appendix.
The created samples are used for training and evaluation in \S~\ref{sec:fine-vs-gen} and \S~\ref{sec:evaluation_other_datasets}. 

The keywords were selected based on dataset statistics. 
For specific applications, a domain expert may have to select other terms (e.g. specific colors in the fashion industry).

\subsection{Training Details}

In our experiments, we use CLIP's image and text encoder from the ViT-B/32 model,\footnote{\href{https://huggingface.co/openai/clip-vit-base-patch32}{https://huggingface.co/openai/clip-vit-base-patch32}} which has 87 million and 63 million parameters respectively. 
And hence, is a relatively small model compared with current multimodal GenAI models.
Our code is modular, and any image and text encoder from \texttt{transformers} \citep{wolf2020transformers} can be used in the framework.

We train all models using a batch size of 64. 
For concepts with multiple negative examples (i.e., \textit{color}, \textit{object}), we train models with 1, 2, and 3 hard negatives. This results in a proportion of hard negatives of 1.5\%, 3\%, and 4.5\%, respectively. We only use one hard negative for the other concepts. 

In all experiments, we use Adam optimizer \citep{kingma2014adam} with a weight decay of $0.1$.
We run evaluations with different learning rates ($\num{5e-7}$, $\num{1e-6}$, $\num{5e-6}$, $\num{1e-5}$, $\num{5e-5}$) and finally use $\num{5e-6}$, since it leads to the best trade-off results for evaluation displayed in \Cref{fig:plots}. 
We do not use a learning rate scheduler since weights are already aligned. 
(This is not the case if encoders that were not trained simultaneously are used.)
Checkpoints are saved every 10\% of the data, and we train for 3 epochs.

The training time for all models was less than two hours on an Nvidia V100. 
Utilizing FP16 training, the GPU memory consumption remained below 12 GB.

\subsection{Training Data}

CLIP is pretrained on 400 million image-text pairs from web sources, and we continue the model pretraining. 
We use the COCO image captioning dataset \citep{lin2015microsoft}, which has 591 thousand image-text pairs (2017 train version).
For each concept, we filtered out samples where at least one keyword was present so that a keyword substitution could be applied.
For evaluation, we use the validation set of COCO 2017.
The respective dataset sizes are shown in the Appendix in \Cref{tab:dataset-size}.

\section{InpaintCOCO: Challenge Set from the Visual Perspective}
\label{sec:challenge_set}
\label{sec:CreateHN-image}

Many multimodal tasks, such as VL Retrieval and Visual Question Answering, present results in terms of overall performance. 
Unfortunately, this approach overlooks more nuanced concepts, leaving us unaware of which specific concepts contribute to the success of current models and which are ignored.
More recent benchmarks attempt to assess particular aspects of vision-language models in response to this limitation. 
Some existing datasets focus on linguistic concepts utilizing one image paired with multiple captions; others adopt a visual or cross-modal perspective.
In this study, we are particularly interested in fine-grained visual concept understanding, which we believe is not covered in existing benchmarks in sufficient isolation. 
Therefore, we create the InpaintCOCO dataset with image pairs with minimum differences that lead to changes in the captions.

\paragraph{Related Work.}

Benchmarks such as ARO \citep{yuksekgonul2022ARO} or VL-CheckList \citep{zhao2022VLchecklist} evaluate models from the language perspective.
ARO examines understanding of attributes and relations without using specific concepts such as color or size.
VL-CheckList\footnote{Parts of the dataset are not available anymore.} is a dataset that investigates concrete concepts such as location, size, material, color, and relations.

On the other side, SVO \citep{hendricks2021SVO} is a dataset that allows analysis from the visual perspective (2 images with 1 caption). Here, relations that deal with verbs are examined.
To our knowledge, the only dataset that deals with fine-grained comprehension from a cross-model perspective is Winoground \citep{thrush2022winoground}.
The dataset consists of two image-text pairs that are very similar to each other. 
This benchmark probes object relations that do not refer to specific concepts. 
The images show similar concepts but are very dissimilar in overall appearance. For samples of the two latter datasets, see \Cref{fig:svo_winoground} in the Appendix.
Both datasets contain real-world images that are in some ways similar in terms of objects, but the scenes still differ significantly. Therefore, it is difficult to tell which image differences cause the model predictions.

We overcome this limitation by creating InpaintCOCO, the first dataset with only minor changes in the \textit{visual} components, so that concept comprehension can be analyzed in a more standardized setting.

\paragraph{Dataset Creation.}

The dataset creation process can be viewed a complement to textual hard-negative samples (\S~\ref{sec:contrastive-learning-with-hn}) in the visual domain. Unlike keyword substitutions, this cannot be done automatically with sufficient accuracy. Even though image segmentation and generative inpainting tools reach impressive results, they still require human supervision to produce high-quality images. Creating a high-quality test set, therefore, requires annotation work.

To generate hard negative image samples, we need to change individual details in the image so that the textual image description no longer fits. 
The procedure is illustrated in \Cref{fig:inpainting}.

The annotation proceeds as follows: The annotators are provided with an image and decide if they want to edit it. If yes, they input the prompt for the object that should be replaced. 
Using the open vocabulary segmentation model CLIPSeg\footnote{\href{https://huggingface.co/CIDAS/clipseg-rd64-refined}{https://huggingface.co/CIDAS/clipseg-rd64-refined}} \cite{Luddecke_2022_CVPR} we obtain a mask for our object of interest (i.e., ``fire hydrant''). Then, the annotator inputs a prompt for Stable Diffusion v2 Inpainting\footnote{\href{https://huggingface.co/stabilityai/stable-diffusion-2-inpainting}{https://huggingface.co/stabilityai/stable-diffusion-2-inpainting}} \cite{Rombach_2022_CVPR} (e.g. with the prompt ``yellow fire hydrant''), which shows three candidate images.
The annotators can try new prompts or skip the current image if the result is insufficient. Finally, the annotator enters a new caption that matches the edited image.
See~\Cref{sec:appendix} for all details.
The images and captions come from the COCO 2017 validation data. We only use images that contain the desired concept and where licenses allow adaptations.

We provide 452 images for the concept \textit{object}, 465 for \textit{color}, and 343 for \textit{size}. 
In contrast to the training process, objects in images are only replaced with objects from the same COCO super category, i.e., ``cat'' with another animal or ``chair'' for another piece of furniture. 
Since \textit{location} would require erasing at one spot and implanting objects at another in a nontrivial way (especially regarding depth), we discard this one concept in the newly created dataset. 
The dataset will be available upon publishing via the HuggingFace hub.

\section{Experiments}

We run several experiments to evaluate the proposed method. We compare the original OpenAI model (\textbf{Orig.}), 
with continued pretraining CLIP model using the classical contrastive learning approach (\textbf{Clas.}) and 
our method with 1 up to 3 hard negative values per batch (\textbf{HN1}, \textbf{HN2}, \textbf{HN3}).
We measure both how well concepts are learned and whether the general image retrieval capability of the model changes on the COCO dataset (\S~\ref{sec:fine-vs-gen}). Additionally, we evaluate the method using our InpaintCOCO challenge set (\S~\ref{sec:dynamic_benchmarking}),
and several other datasets (\S~\ref{sec:evaluation_other_datasets}).

\subsection{Fine-grained vs. Coarse Understanding}
\label{sec:fine-vs-gen}

\paragraph{Fine-grained Concept Understanding.}
Here, we are interested in whether models have learned detailed concept knowledge. 
We pose the evaluation as a ranking problem with one image on one side and $n$ different texts on the other side. 
Besides the correct text (containing the correct keyword), we generate all possible $n-1$ negative examples using the same procedure as in \S~\ref{sec:CreateHN} and rank the texts with the model. We evaluate the ranking using the top-1 accuracy.
See \Cref{tab:cc_example} in the Appendix for some examples.

\paragraph{General Image Retrieval.} 
We evaluate the general capabilities of our models using the COCO dataset.
We want the general retrieval capability to remain high even though we train our models with a focus on one concept. 
We report text-to-image retrieval Recall@5 on the whole COCO validation set.

\begin{figure*}[!h]
\centering
\begin{subfigure}[t]{0.48\textwidth}
\centering
\includegraphics[width=.999\textwidth]{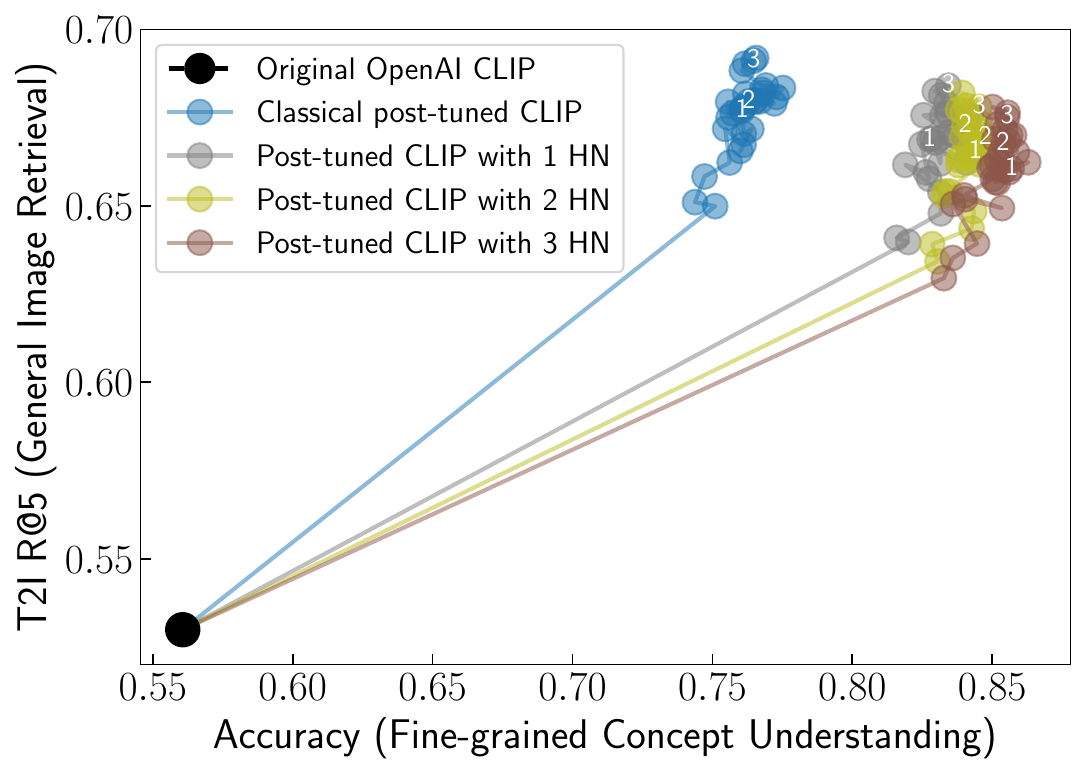}
\caption{Object}
\label{fig:Object}
\end{subfigure}%
\hfill
\begin{subfigure}[t]{0.48\textwidth}
\centering
\includegraphics[width=.999\textwidth]{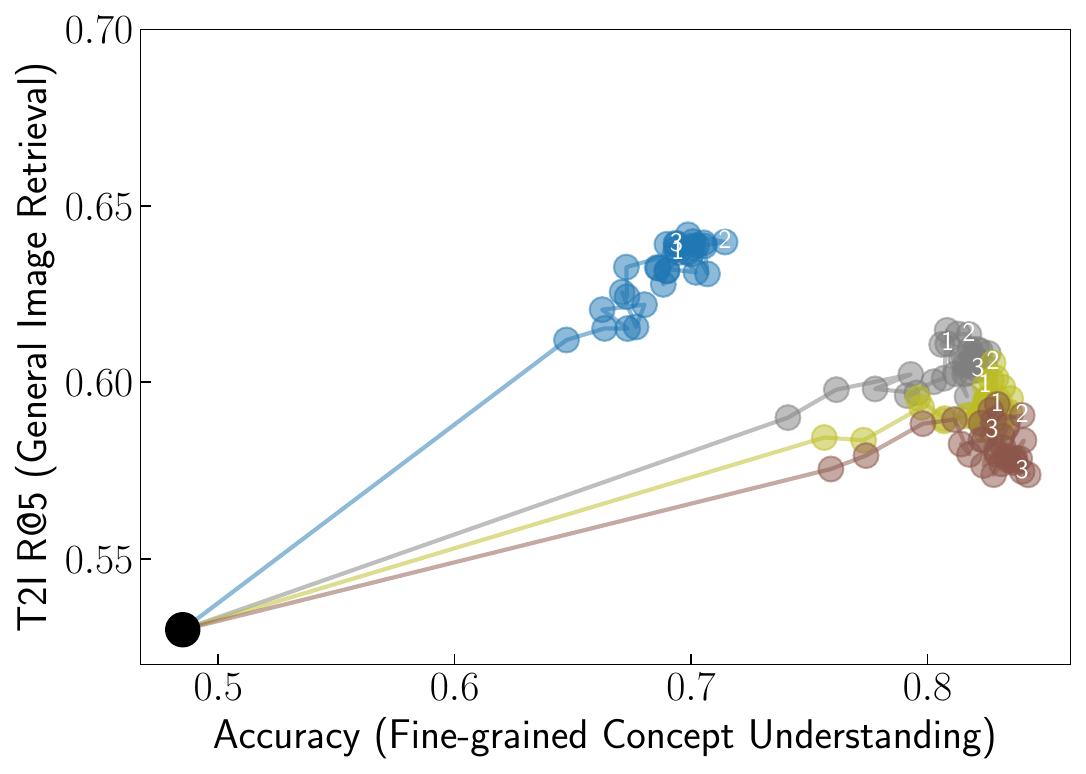}
\caption{Color}
\label{fig:Color}
\end{subfigure}


\begin{subfigure}[t]{0.48\textwidth}
\centering
\includegraphics[width=.999\textwidth]{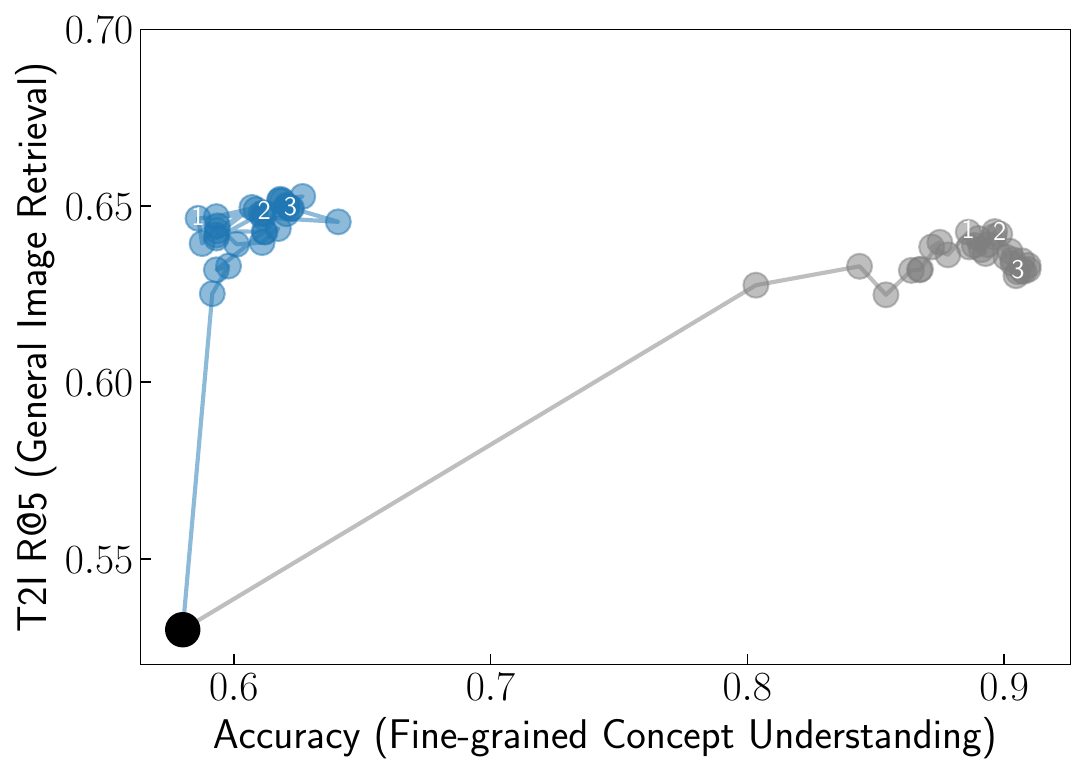}
\caption{Location}
\label{fig:Location}
\end{subfigure}%
\hfill
\begin{subfigure}[t]{0.48\textwidth}
\centering
\includegraphics[width=.999\textwidth]{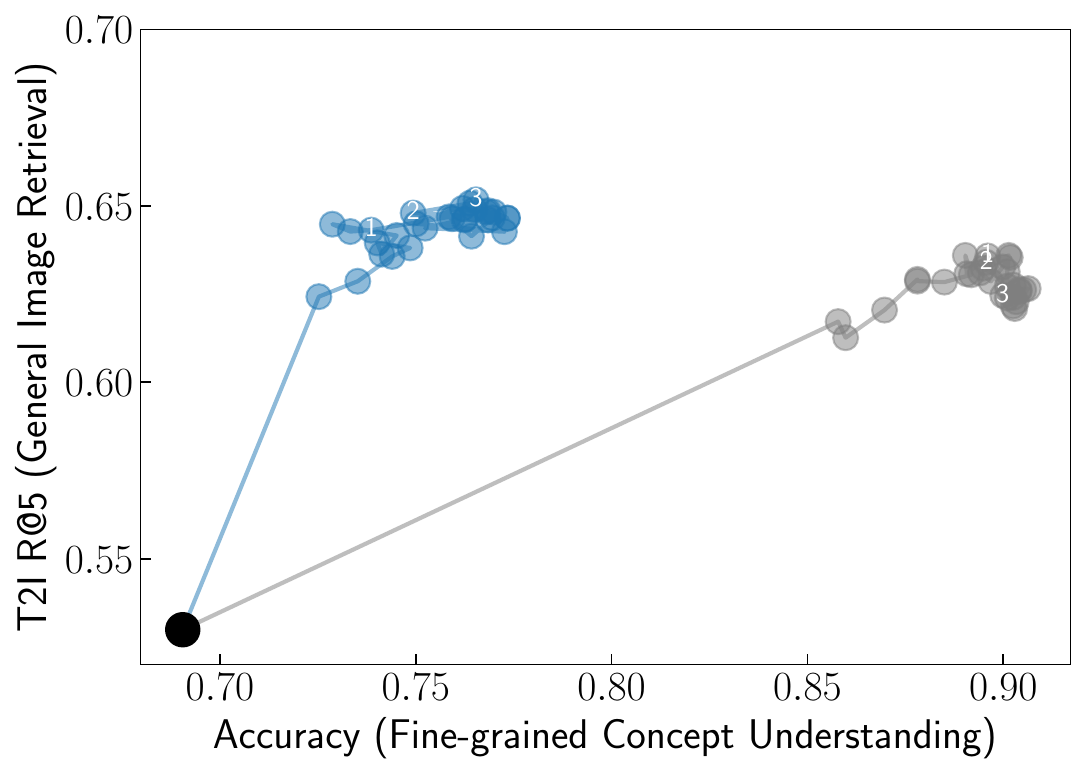}
\caption{Size}
\label{fig:Size}
\end{subfigure}

\caption{Fine-grained Concept Understanding vs. General Image Retrieval: Results for four different concepts trained on corresponding dataset subsets. Checkpoints are evaluated after every 10\% of the data (circles); checkpoints at epoch ends are marked with the respective numbers. The results are also in a table form in Table~\ref{tab:results} in the Appendix.} 
\label{fig:plots}
\end{figure*}

\paragraph{Results.}
Results are shown in \Cref{fig:plots} (and exact results per epoch are displayed in \Cref{tab:results} in the Appendix).
The performance of the original OpenAI CLIP is shown with a black dot and constantly reaches the worst results.
Continued pretraining on COCO massively increases the general retrieval performance for all concepts, showing the successful domain adaptation regarding this dataset.
It also improves the concept understanding for all concepts except for \textit{location}.
We are especially interested in the trade-off between general retrieval and concept understanding for the different types of further pretrained models. 
The following values are based on model checkpoints at the end of the epoch. 

For \textit{\textbf{objects}}, we observe a performance increase regarding the fine-grained comprehension from 0.56 for the original OpenAI CLIP models to 0.76 when further pretraining on COCO. 
Using the hard negatives approach, performance increases by 7 to 10 percentage points, depending on the hard negative proportion and training duration. 
Here, the general retrieval performance only drops by 1 to 2 percentage points in relation to the classical approach.

We observe a similar pattern can be seen for the concept \textit{\textbf{color}}.
Understanding of the concept improved by 11 to 15 percentage points compared to the original training process. 
On the other hand, the general comprehension pattern loses 3 to 4 percentage points with HN1 and slightly more with hard negative values. 
In this case, one hard example seems sufficient to learn concepts.

A single hard negative sample per batch is enough to enhance spatial understanding (\textit{\textbf{location}}). 
Here, concept comprehension improves by 29 or 30 percentage points (depending on the duration of training) to around 90\%. 
Both the original CLIP model and the further pretrained model only achieved around 60\%.
The general understanding decreases by 1 percentage point only or by 2 percentage points with long training.
A substantial improvement in location comprehension is achieved with negligible loss in overall understanding

We observe a similar pattern for the \textit{\textbf{size}} concept. 
Further pretraining on COCO improves concept understanding by 5 to 8 percentage points to around 75\%. 
On the other hand, if the new learning concept is used, the result is 90\%, which corresponds to an additional improvement of 15 percentage points. 
Depending on training time, the general image retrieval capabilities lose no or only 3 percentage points. 
As with location, an almost continuous improvement can be observed. 

Across all concepts, hard negative contrastive learning can significantly increase concept understanding. 
This also applies to settings with just one hard example. 
It is shown that with a small adjustment in the objective, the models can learn a much more complex understanding of image and text. 
Meanwhile, the general ability to represent images and texts is hardly affected. 

\begin{table}
\small
\centering
\begin{subtable}[t]{0.48\textwidth}
\centering
\begin{tabular}{l|rrrrr}
\toprule
Epoch &  Orig. &  Clas. &  HN1 &  HN2 &  HN3 \\  \midrule
1     &   \oC{0.78} &   \oC{0.85} & \oC{0.88} & \oC{0.88} & \oC{0.88} \\
2     &   \oC{0.78} &   \oC{0.84} & \oC{0.86} & \oC{0.88} & \oC{0.88} \\
3     &   \oC{0.78} &   \oC{0.87} & \oC{0.88} & \oC{0.89} & \oC{0.90} \\ \bottomrule
\end{tabular}
\caption{Object}
\label{tab:cs_object}
\end{subtable}%

\bigskip 

\begin{subtable}[t]{0.48\textwidth}
\small
\centering
\begin{tabular}{l|rrrrr}
\toprule
Epoch &  Orig. &  Clas. &  HN1 &  HN2 &  HN3 \\ \midrule
1     &   \cC{0.62} &   \cC{0.83} & \cC{0.89} & \cC{0.89} & \cC{0.89} \\
2     &   \cC{0.62} &   \cC{0.83} & \cC{0.90} & \cC{0.90} & \cC{0.91} \\
3     &   \cC{0.62} &   \cC{0.83} & \cC{0.90} & \cC{0.91} & \cC{0.92} \\ \bottomrule
\end{tabular}
\caption{Color}
\label{tab:cs_color}
\end{subtable}%

\bigskip 

\begin{subtable}[t]{0.48\textwidth}
\small
\centering
\begin{tabular}{l|rrr}
\toprule
Epoch &  Orig. &  Clas. &  HN1  \\ \midrule
1     &   \sC{0.26} &   \sC{0.27} & \sC{0.55} \\
2     &   \sC{0.26} &   \sC{0.28} & \sC{0.57} \\
3     &   \sC{0.26} &   \sC{0.30} & \sC{0.60} \\ \bottomrule
\end{tabular}
\caption{Size}
\label{tab:cs_size}

\end{subtable}%

\caption{Accuracy for fine-grained understanding from the visual perspective following \Cref{eq:challengeset} for InpaintCOCO dataset.}
\label{fig:challenge_set}
\end{table}

\subsection{Challenge Set Results}
\label{sec:dynamic_benchmarking}

The ranking-based evaluation in the previous section only assessed the model capabilities from the language perspective in a very similar setup to how the model was trained. This section assesses the model from the vision perspective using our InpaintCOCO challenge set.

We consider an image pair from the InpaintCOCO dataset correctly classified if the correct images would be more likely to be retrieved based on the original caption and the newly created caption.
This leads to the formula,

\begin{equation}
\begin{split}
\operatorname{sim}(i_\text{COCO},t_\text{COCO}) > \operatorname{sim}(i_\text{inp},t_\text{COCO}) \quad \land \\
\operatorname{sim}(i_\text{inp},t_\text{inp}) > \operatorname{sim}(i_\text{COCO},t_\text{inp})
\end{split}
\label{eq:challengeset}
\end{equation}
\noindent with image $i$ and text $t$, originating from the original COCO and InpaintCOCO dataset.
The corresponding results for each concept are displayed in \Cref{fig:challenge_set}.


The results show that continued pretraining improves understanding of all three concepts (object, color, size). 
The improvements between the original OpenAI CLIP model and the further pretrained model are 6 to 9 percentage points for the object and 21 for the color concept.
The improvements are less distinct for size, with a 1 to 4 percentage point gain, which aligns with the textual comprehension results displayed in \Cref{fig:Size}.


For the \textit{\textbf{object}} concept, hard negative training brought a 7 to 10 percentage point improvement for the textual viewpoint (see \Cref{tab:results}). 
For the evaluation from the visual perspective, improvements are 2 to 4 percentage points.
This relatively smaller improvement is likely because each object could be replaced with the 79 other object names during training. 
Still in this evaluation, a replacement was only executed within the COCO super-category (5 to 10 object names per super-category). 
The differences in performance between the three models using hard negative training is $\leq 2$ percentage points.

The evaluation of the \textit{\textbf{color}} concept shows an improvement of 6 to 9 percentage points for the visual perspective. 
From the textual perspective (\Cref{tab:results}), the improvement was at over 11 percentage points. 
As before, using more hard negative samples during training does not further improve the performance systematically ($\leq 2$ percentage point).

For the \textit{\textbf{size}} concept, we see a big improvement for both perspectives when using hard negative models. 
From the visual perspective, there is an improvement of over 28 percentage points (\Cref{tab:cs_size}), and from the textual perspective, 
13 to 16 percentage points.


The results show that training for just one epoch is sufficient for learning the concepts \textit{object}, \textit{color}, and  \textit{size}, and further training does not continue improving the results systematically. 
Additionally, using a single hard negative is sufficient to improve understanding of the concepts.

\begin{table*}[!h]
\small
\centering
\begin{tabular}{@{}llr|rrrrr@{}}
\toprule
Dataset & Concept & Count & Orig. & Clas. & HN1 & HN2 & HN3 \\   \midrule
Flickr30k & object & 30,926 & \oCC{.48} & \oCC{.66} & \oCC{.75} & \oCC{.77} & \oCC{.79} \\ 
  & color & 45,003 & \cCC{.50} & \cCC{.64} & \cCC{.72} & \cCC{.73} & \cCC{.74}    \\ 
  & location & 20,097 & \lCC{.62} & \lCC{.64} & \lCC{.89} &  &  \\ 
  & size & 14,853 & \sCC{.75} & \sCC{.79} & \sCC{.92} &  &  \\ \midrule
SBU & object & 200,310 & \oCC{.33} & \oCC{.41} & \oCC{.48} & \oCC{.49} & \oCC{.50}   \\ 
              & color & 162,652 & \cCC{.51} & \cCC{.54} & \cCC{.61} & \cCC{.62} & \cCC{.62} \\
              & location & 159,673 & \lCC{.61} & \lCC{.60} & \lCC{.79} &  &  \\
              & size & 47,069 & \sCC{.63} & \sCC{.65} & \sCC{.73} &  &  \\ \midrule
Fashion200K & object & 3,628 & \oCC{.24} & \oCC{.25} & \oCC{.35} & \oCC{.38} & \oCC{.39}   \\
            & color & 141,413 & \cCC{.68} & \cCC{.68} & \cCC{.69} & \cCC{.69} & \cCC{.70} \\   \midrule
NASA Earth Instagram  & color & 132 & \cCC{.43} & \cCC{.48} & \cCC{.55} & \cCC{.55} & \cCC{.58} \\ \midrule
Old Book Illustrations & object & 124 &  \oCC{.27} & \oCC{.35} & \oCC{.38} & \oCC{.35} & \oCC{.31}   \\ 
              & location & 95 & \lCC{.61} & \lCC{.50} & \lCC{.77} &  &  \\
              & size & 82 & \sCC{.61} & \sCC{.63} & \sCC{.79} &  &  \\ \bottomrule
\end{tabular}
\caption{Fine-grained concept understanding results (accuracy) for a diverse selection of datasets where the sample size is larger $50$. Evaluated on dataset subsets where corresponding keywords are present.}
\label{tab:other_datasets}
\end{table*}

\subsection{Evaluations on other Datasets}
\label{sec:evaluation_other_datasets}

We further investigate the performance of our model using more VL datasets.
First, we evaluate the models using general VL datasets. The investigated concepts occur with different frequencies, and for high-quality results, it is important that these concepts are understood to increase the overall performance. Therefore, we further investigate fine-grained concept understanding on the Flickr30k \citep{young2014image} and SBU Captioned Photo \citep{Ordonez2011SBU} datasets. 

Fine-grained concept understanding is also important in specific domains. For example, in fashion, a correct assignment of garments and colors is important, not the mere presence of colors in the image. For this analysis, we evaluated our models on the very specific datasets Fashion200K \citep{han2017fashion200k}, NASA Earth Instagram,\footnote{\href{https://huggingface.co/datasets/nkasmanoff/nasa\_earth\_instagram}{https://huggingface.co/datasets/nkasmanoff/nasa\_earth\_ instagram}} and 
Old Book Illustrations.\footnote{\href{https://huggingface.co/datasets/gigant/oldbookillustrations}{https://huggingface.co/datasets/gigant/oldbookillustrations}} These datasets are very heterogeneous in their appearance.

All models achieve good results except the \textit{color} concept on the Fashion200k dataset and \textit{object} concept on Old Book Illustrations. 
For the former, this is because images usually show garments with a distinct color. Yet, there is little background noise or noise from irrelevant items, which can confuse the color alignment of the model in this dataset. 
The latter shows old-fashioned drawings with objects very dissimilar to those in the COCO dataset.
Our approach to learn concepts works very well for the remaining evaluations.

\section{Conclusion}

We introduce a robust method for enhancing fine-grained concept understanding with minimal impact on general retrieval capabilities using hard negative sampling in contrastive learning.
We show that various concepts can be learned efficiently with minor text input adjustments.
Moreover, improvements in concept understanding are observable after continued pretraining on only 10\% of our data.
Furthermore, one hard negative sample per image in a batch of 64 proves sufficient to incorporate the concept of interest into the model.

We comprehensively evaluate our method on several datasets, including our new challenge set. 
Our method outperforms classical contrastive learning on all investigated concepts.
Existing datasets often focus on linguistic perturbations or use dissimilar images, precluding a structured evaluation of permuted visual concepts in isolation. 
To address this gap, InpaintCOCO represents the first dataset adjusting minor image parts in a controlled setting, facilitating cross-model fine-grained understanding. 
This ensures that the model's output is influenced only by one object and not the rest of the scene.

The results show that fine-grained concept understanding also generalizes to images of different styles when using InpaintCOCO and domain-specific datasets. 
Our method is data-efficient and requires only a little domain knowledge to design the hard negatives. 
This makes it particularly suitable for domain adaptation in image retrieval, as well as for developing new CLIP-based models, e.g., for object detection \citep{minderer2023scaling}.

\section*{Acknowledgements}

The authors gratefully acknowledge the compute time granted by Monacum One at Bundeswehr University Munich. The work at CUNI, including annotation payments, was supported by Charles University project PRIMUS/23/SCI/023.


\section*{Limitations and Risks}

Our research introduces a novel method for training CLIP aimed at incorporating concepts and representations that are challenging to learn using current approaches. In our experiments, we only used one specific CLIP model; however, we believe there is no reason why the method should not work or work systematically differently for smaller or larger CLIP models.

We conducted training with the well-studied COCO captioning dataset, which is standard in multimodal research. The proposed method is expected to show consistent performance also using other multimodal training datasets. 
Notably, evaluations on out-of-domain datasets, where the model was not trained, emphasize the robustness of our approach.
One essential prerequisite for our methodology to work is the presence of keywords of interest in the training corpus and language and domain knowledge to decide how the keywords should be replaced. 
The keyword substitution will be more difficult in languages with more complex morphology than in English.
Experiments involved using four concepts with a carefully chosen set of keywords.
Depending on domain-specific tasks, other keywords might be of interest (e.g., a large list of garments for the fashion domain).

COCO is a dataset with image-text pairs where the captions are proper sentences, displaying a specific level of detail, and are carefully created by annotators. The images in the COCO dataset come from Flickr from 2014; therefore, they reflect the Flickr user structure at that time, i.e., the images mostly show the Western world and/or other countries from the Western perspective. The captions are in English. Thus, the model we developed does not generalize well beyond the Western world. However, we believe that is the limitation of the dataset, and the presented method itself is dataset agnostic.

The primary application goal of the models we worked with is to make image collections better accessible. Similar to other work on this VL modeling that enables better image understanding at scale, there is a risk of using technology based on the models for activities such as large-scale video surveillance.

\bibliography{anthology,custom}

\appendix
\section{Appendix}
\label{sec:appendix}

\paragraph{Creating Hard Negative Samples}
In \Cref{tab:heuristic-permutation}, we specify all used substitution keywords for the concepts \textit{object}, \textit{color}, \textit{location}, and \textit{size}.

\Cref{tab:cc_example} lists text samples for the fine-grained concept understanding task, which is used in \S~\ref{sec:fine-vs-gen} for each concept.

\begin{table}[h!]
\small
\centering
\begin{tabular}{p{1cm}p{5.75cm}}
\toprule
Concept  & Keywords \\ \midrule 
object & [person, bicycle, car, motorbike, aeroplane, bus, train, truck, boat, traffic light, fire hydrant, stop sign, parking meter, bench, bird, cat, dog, horse, sheep, cow, elephant, bear, zebra, giraffe, backpack, umbrella, handbag, tie, suitcase, frisbee, skis, snowboard, sports ball, kite, baseball bat, baseball glove, skateboard, surfboard, tennis racket, bottle, wine glass, cup, fork, knife, spoon, bowl, banana, apple, sandwich, orange, broccoli, carrot, hot dog, pizza, donut, cake, chair, sofa, potted plant, bed, dining table, toilet, tv monitor, laptop, mouse, remote, keyboard, cell phone, microwave, oven, toaster, sink, refrigerator, book, clock, vase, scissors, teddy bear, hair drier, toothbrush] \\ 
color &  [blue, red, green, yellow, black, white, brown, gray, orange] \\
location &       left  $\leftrightarrow$ right, 
     above  $\leftrightarrow$  below, 
     under  $\leftrightarrow$  over, 
     foreground  $\leftrightarrow$  background,
     in front of  $\leftrightarrow$  behind,
     back  $\leftrightarrow$  front  \\
size  &  large $\leftrightarrow$   small,  
 little $\leftrightarrow$   big,  tall $\leftrightarrow$   short,   
 long $\rightarrow$   short,
 thin $\leftrightarrow$   fat, 
 huge $\leftrightarrow$   tiny,
  giant $\rightarrow$   tiny\\  \bottomrule
\end{tabular}
\caption{All keywords that can be replaced for the four concepts. That are 80 for \textit{object}, 9 for \textit{color}, 12 for \textit{location} and 11 for \textit{size}. In lists, each word can be replaced by any other word. ``$\leftrightarrow$'' and  ``$\rightarrow$'' denote words that can be replaced in both and one direction, respectively.}
\label{tab:heuristic-permutation}
\end{table}

\begin{table}[!h]
\small
\centering
\begin{subtable}[t]{0.48\textwidth}
\centering
\begin{tabular}{p{0.05cm}p{6.7cm}}
\toprule
\xmark & A small yellow person on a branch of a tree. \\ 
\xmark & A small yellow bicycle on a branch of a tree. \\ 
\xmark & ... \\ 
\checkmark & A small yellow \textbf{bird} on a branch of a tree. \\ 
\xmark & A small yellow cat on a branch of a tree. \\ 
\xmark & ... \\  \bottomrule
\end{tabular}
\caption{Object (for all 80 object names)}
\label{tab:cc_example_object}
\end{subtable}%

\bigskip 

\begin{subtable}[t]{0.48\textwidth}
\small
\centering
\begin{tabular}{p{0.05cm}p{6.7cm}}
\toprule
\xmark & a blue cat sits under a black open umbrella. \\
\xmark & a red cat sits under a black open umbrella. \\
\xmark & a green cat sits under a black open umbrella. \\
\xmark & a yellow cat sits under a black open umbrella. \\
\xmark & a black cat sits under a black open umbrella. \\
\checkmark & a \textbf{white} cat sits under a black open umbrella. \\
\xmark & a brown cat sits under a black open umbrella. \\
\xmark & a gray cat sits under a black open umbrella. \\
\xmark & a orange cat sits under a black open umbrella. \\    \bottomrule
\end{tabular}
\caption{Color}
\label{tab:cc_example_color}
\end{subtable}%

\bigskip 

\begin{subtable}[t]{0.48\textwidth}
\centering
\small
\begin{tabular}{p{0.05cm}p{6.7cm}}
\toprule
\checkmark & a white cat sits \textbf{under} a black open umbrella. \\
\xmark & a white cat sits over a black open umbrella. \\    \bottomrule
\end{tabular}
\caption{Location}
\label{tab:cc_example_location}
\end{subtable}%

\bigskip 

\begin{subtable}[t]{0.48\textwidth}
\centering
\small
\begin{tabular}{p{0.05cm}p{6.7cm}}
\toprule
\checkmark & A \textbf{small} yellow bird on a branch of a tree. \\ 
\xmark & A large yellow bird on a branch of a tree. \\   \bottomrule
\end{tabular}
\caption{Size}
\label{tab:cc_example_size}
\end{subtable}%
\caption{Exemplary image descriptions that are used as text samples in the fine-grained concept understanding task (see \S~\ref{sec:fine-vs-gen}) for the example images displayed in \Cref{fig:examples} or \Cref{fig:inpainting}.}
\label{tab:cc_example}
\end{table}

\paragraph{COCO training and evaluation data.}

\Cref{tab:dataset-size} shows the size of the training datasets and fine-grained concept understanding evaluation datasets.
The images predominantly depict scenes from the USA and Western countries, and all captions are exclusively in English.

There are many large-scale VL datasets available to pretrain or further pretrain models (e.g., Conceptual Caption \citep{sharma2018conceptual} or LAION-5B \citep{schuhmann2022laionb} with 3.3 million and 5 billion image-text pairs respectively).
Yet, results in \Cref{fig:plots} indicate that training on a small-sized COCO dataset (even for just one epoch) is sufficient to learn the concepts of interest.

\begin{table}[h!]
\centering
\small
\begin{tabular}{@{}lrr@{}}
\toprule
Concept & Further pretraining & Evaluation \\ 
 & dataset size &  dataset size\\ \midrule
object & 305,056 & 12,907 \\
color & 92,006 & 3,907 \\
location & 58,156 & 2,527 \\
size & 60,626 & 2,601 \\  \bottomrule
\end{tabular}
\caption{COCO dataset size for all concepts used for further pretraining CLIP and evaluation.}
\label{tab:dataset-size}
\end{table}

\begin{figure}[!h]
\begin{subfigure}[t]{0.48\textwidth}
\centering
\small
\begin{tabular}{p{3cm}p{3cm}}
\toprule
\includegraphics[width=3cm,keepaspectratio]{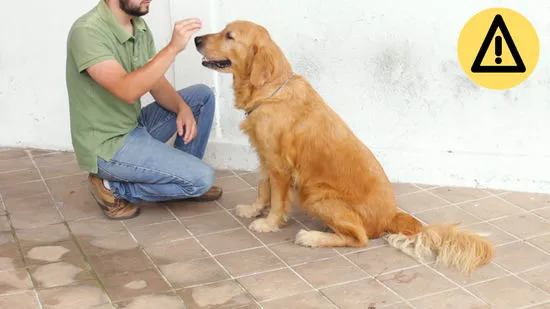} & \includegraphics[width=3cm,keepaspectratio]{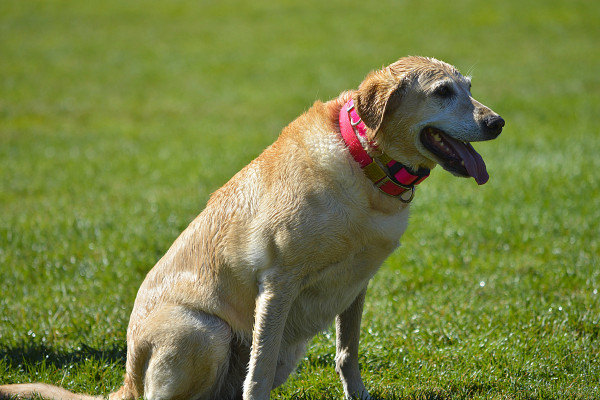}  \\
A dog is sitting on the floor. &  \\  \midrule
\includegraphics[width=3cm,keepaspectratio]{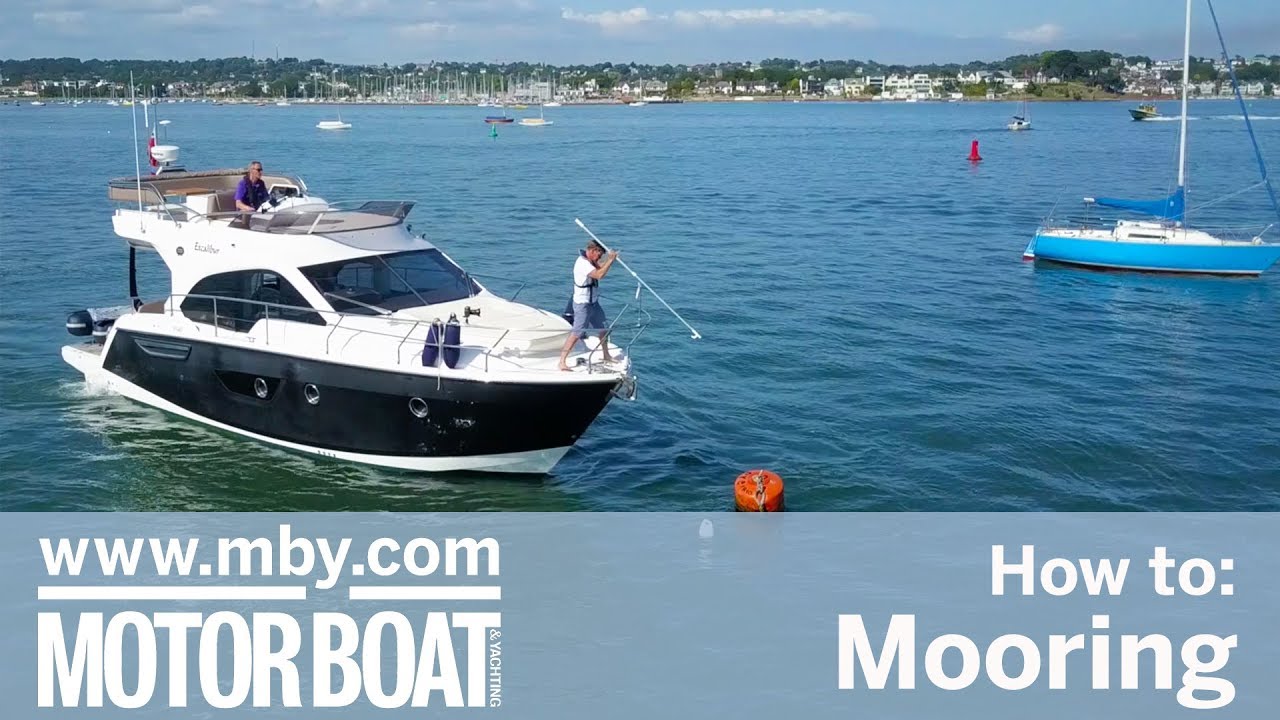} & \includegraphics[width=3cm,keepaspectratio]{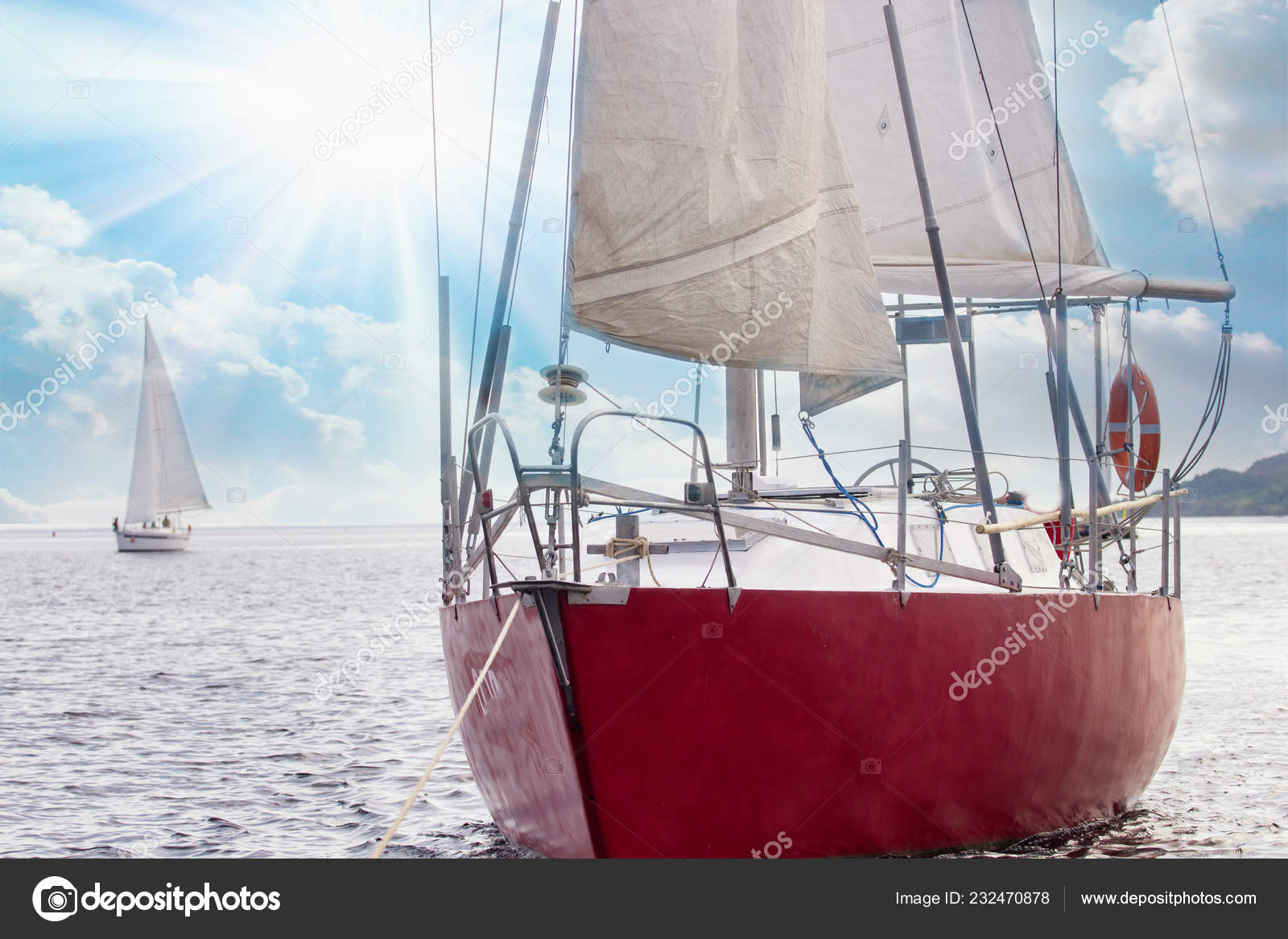}  \\ 
A boat can moor while at sea. &   \\ \midrule  
\end{tabular}
\caption{SVO-Probes}
\label{tab:svo}
\end{subfigure}%

\bigskip 

\begin{subfigure}[t]{0.48\textwidth}
\centering
\small
\begin{tabular}{p{3cm}p{3cm}}
\toprule
\includegraphics[width=3cm,keepaspectratio]{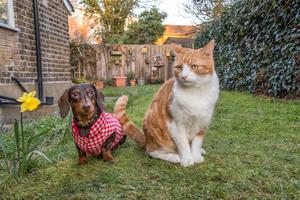} & \includegraphics[width=3cm,keepaspectratio]{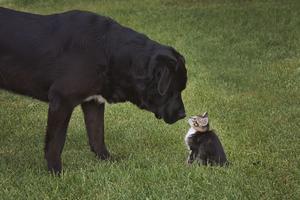}  \\
a big cat is next to a small dog & a small cat is next to a big dog \\ \midrule
\includegraphics[width=3cm,keepaspectratio]{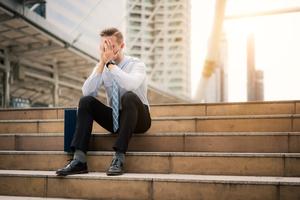} & \includegraphics[width=3cm,keepaspectratio]{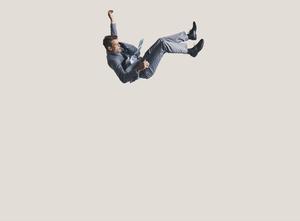}  \\ 
the businessperson's down fall & the businessperson's fall down  \\ \midrule
\end{tabular}
\caption{Winoground}
\label{tab:winoground}
\end{subfigure}%

\caption{Samples from visual and cross-model datasets.}
\label{fig:svo_winoground}
\end{figure}

\paragraph{Fine-grained VL Benchmarks.}
In \Cref{tab:dataset_list}, we list different benchmark datasets for fine-grained understanding in VL. 
Some image-text samples for ARO and Winoground are displayed in \Cref{fig:svo_winoground} since these datasets provide two images per sample -- similar to InpaintCOCO. 
However, unlike InpaintCOCO, the visual representations are very different regarding the scenes presented in these datasets.

\begin{table*}[th]
\centering
\small
\begin{tabular}{@{}lllll@{}}
\toprule
Dataset & Size & Perspective & Samples & Tasks \\  \midrule
ARO & 77k & Linguistic & 1 image $\leftrightarrow$ 2 texts & Attributes, Relations, Order understanding \\
VL-CheckList & 410k & Linguistic & 1 image $\leftrightarrow$ 2 texts & Attributes, Relations, Object understanding \\
SVO-Probs & 48k & Visual & 2 images $\leftrightarrow$ 1 text & Relations (Verb Understanding) \\
Winoground & 400 & Cross-modal & 2 images $\leftrightarrow$ 2 texts & Relations \\
InpaintCOCO & 1,260 & Cross-modal & 2 images $\leftrightarrow$ 2 texts & Attributes, Object understanding \\  \bottomrule
\end{tabular}
\caption{Datasets for fine-grained understanding in VL.}
\label{tab:dataset_list}
\end{table*}

\begin{figure}[]
\includegraphics[clip, trim=4.5cm 1.5cm 4.5cm 1.9cm, , width=.48\textwidth,keepaspectratio]{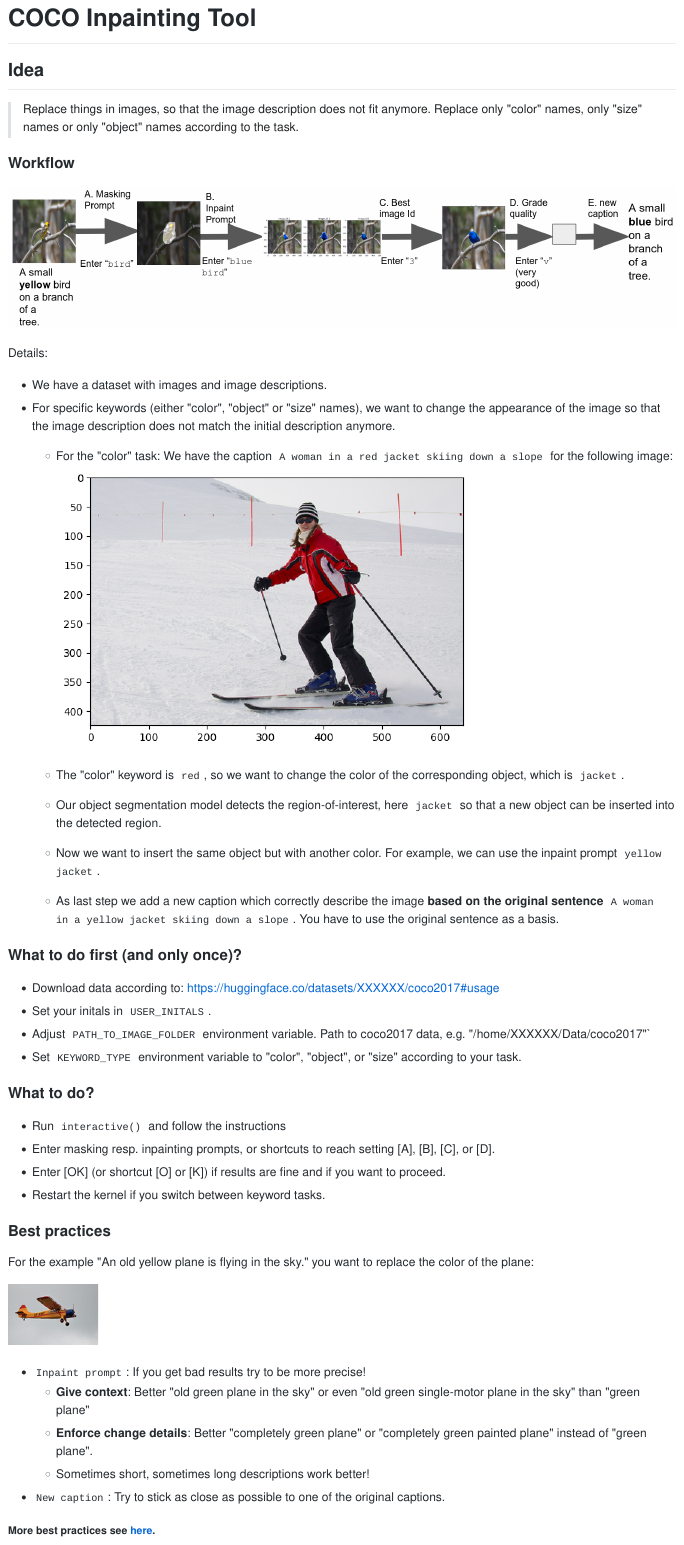} 
\caption{Instructions of inpainting tool provided to student workers.}
\label{tab:instructions}
\end{figure}

\paragraph{Challenge Set Creation.}
Undergraduate student workers created the challenge set.
They were provided with an interactive Python environment with which they interacted via various prompts and inputs.
The description of the task and the problem of the research question was made available to them (see \Cref{tab:instructions}).
In addition to a detailed written explanation of how the tool works, they were also given ``best practices,'' which were created by one student and reviewed by the authors.

For color, any other \textit{color} and for \textit{size}, the opposite statement can be chosen. Yet, within \textit{objects}, the students were asked to replace with objects from the same COCO super-category (to ensure that no ``plain'' needs to be inpainted in an indoor scene). There are 12 super categories for the 80 object names: person, vehicle, outdoor, animal, accessory, sports, kitchen, food, furniture, electronic, appliance, and indoor.

The workflow comprises these steps:
\begin{enumerate}[noitemsep,nolistsep]
    \item A random COCO (2017 validation) image is shown with all its captions containing a concept keyword. 
    \item The annotator enters a masking prompt for the segmentation task based on the object of interest (e.g., ``fire hydrant'' in \Cref{fig:inpainting}). They can also enlarge the mask within the $x$ and $y$ dimensions by passing additional parameters. This is useful if a larger object is to be inserted into the image. Several attempts can be made until the mask meets the requirements. Only then the next step is carried out.
    \item Then, the annotator enters an inpainting prompt (the image generation takes roughly 1 minute). They are provided with three different inpainted images. They proceed if at least one high-quality image has been generated.
    \item The best image is chosen from the three proposals. 
    \item Based on the selection before, they rate the pictures as ``very good'' or ``okay''.
    \item Finally, a new, correct caption is added based on one of the original COCO captions. 
\end{enumerate}

A subset of students had pre-existing roles within the university, while others were purposefully recruited for the designated task. The compensation for student assistants adhered to the legally stipulated wages in their respective countries, amounting to CZK 300.00 per hour in the Czech Republic and EUR 12.00 per hour in Germany.

\paragraph{Challenge Set Details.}

Our InpaintCOCO challenge set is based on the famous COCO dataset.
All captions follow ``Creative Commons Attribution 4.0 License'' 
and hence can be changed. 
Images originate from Flickr, and have diverse licenses (``Attribution License'',
       ``Attribution-NonCommercial-ShareAlike License'',
       ``Attribution-NonCommercial License'',
       ``Attribution-ShareAlike License'') which all allow scientific usage and modification (like inpaining).
Individual licenses are listed in each sample of the dataset.

\paragraph{Experiment Results.}
For numeric results from \Cref{fig:plots} see \Cref{tab:results}. In this Table fine-grained concept understanding results (Accuracy) and COCO text-to-image retrieval results (T2I R@5) are presented.  
``Orig.'', ``Clas.'', ``HN1'', ``HN2'', and ``HN3'' indicate the original \mbox{OpenAI} CLIP model, the classical further pretrained model, and models trained with 1, 2, or 3 hard negative samples.

\begin{table*}[t]
\centering
\small
\begin{tabular}{lr|rrrrr|rrrrr}
\toprule
&  & \multicolumn{5}{c|}{Accuracy} & \multicolumn{5}{c}{T2I R@5}  \\
Concept & Epoch & Orig. & Clas. & HN1 & HN2 & HN3 &  Orig. & Clas. & HN1 & HN2 & HN3  \\ \midrule 
\multirow[t]{3}{*}{object}
 &  1 & \oAc{.56} & \oAc{.76} & \oAc{.83} & \oAc{.84} & \oAc{.86} & \oR{.53} & \oR{.68} & \oR{.67} & \oR{.67} & \oR{.66} \\
 &  2 & \oAc{.56} & \oAc{.76} & \oAc{.84} & \oAc{.85} & \oAc{.85} & \oR{.53} & \oR{.68} & \oR{.67} & \oR{.67} & \oR{.67} \\
 &  3 & \oAc{.56} & \oAc{.76} & \oAc{.83} & \oAc{.85} & \oAc{.86} & \oR{.53} & \oR{.69} & \oR{.68} & \oR{.68} & \oR{.68}  \\ \midrule
\multirow[t]{3}{*}{color}
 &  1 & \cAc{.48} & \cAc{.69} & \cAc{.81} & \cAc{.82} & \cAc{.83} & \cR{.53} & \cR{.64} & \cR{.61} & \cR{.60} & \cR{.59} \\
 &  2 & \cAc{.48} & \cAc{.71} & \cAc{.82} & \cAc{.83} & \cAc{.84} & \cR{.53} & \cR{.64} & \cR{.61} & \cR{.61} & \cR{.59} \\
 &  3 & \cAc{.48} & \cAc{.69} & \cAc{.82} & \cAc{.83} & \cAc{.84} & \cR{.53} & \cR{.64} & \cR{.60} & \cR{.59} & \cR{.57} \\ \midrule
\multirow[t]{3}{*}{location}
 &  1 & \lAc{.58} & \lAc{.59} & \lAc{.89} &    &    & \lR{.53} & \lR{.65} & \lR{.64} &    &     \\
 &  2 & \lAc{.58} & \lAc{.61} & \lAc{.90} &    &    & \lR{.53} & \lR{.65} & \lR{.64} &    &   \\
 &  3 & \lAc{.58} & \lAc{.62} & \lAc{.91} &    &    & \lR{.53} & \lR{.65} & \lR{.63} &    &   \\ \bottomrule
 \multirow[t]{3}{*}{size}
 &  1 & \sAc{.69}  & \sAc{.74} & \sAc{.90} &    &    & \sR{.53} & \sR{.64} & \sR{.64} &    &      \\
 &  2 & \sAc{.69}  & \sAc{.75} & \sAc{.90} &    &    & \sR{.53} & \sR{.65} & \sR{.63} &    &   \\
 &  3 & \sAc{.69}  & \sAc{.77} & \sAc{.90} &    &    & \sR{.53} & \sR{.65} & \sR{.62} &    &     \\ \midrule
\end{tabular}
\caption{Accuracy of fine-grained concept understanding (evaluated on dataset subsets) and COCO text-to-image Recall@5 for general image retrieval (evaluated on whole dataset) for models trained on respectively concepts. Checkpoints for epochs 1 to 3.}
\label{tab:results}
\end{table*}

\begin{figure}[h]
\begin{subfigure}[t]{0.48\textwidth}
\centering
\begin{tabular}{p{3cm}p{3cm}}
\toprule
\includegraphics[width=3cm,keepaspectratio]{assets/inpaintCOCO/img_1_image.jpg} & \includegraphics[width=3cm,keepaspectratio]{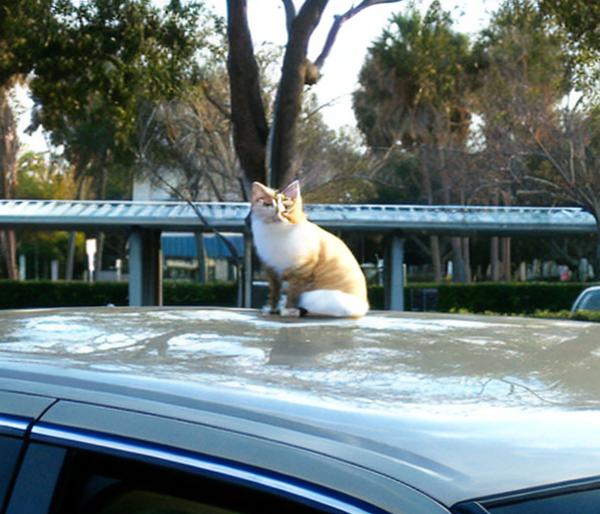}  \\
A \textbf{bird} is standing on top of a car. & A \textbf{cat} is standing on top of a car. \\ \bottomrule
\end{tabular}
\caption{Object}
\label{tab:object}
\end{subfigure}%

\bigskip 

\begin{subfigure}[t]{0.48\textwidth}
\centering
\begin{tabular}{p{3cm}p{3cm}}
\toprule
\includegraphics[width=3cm,keepaspectratio]{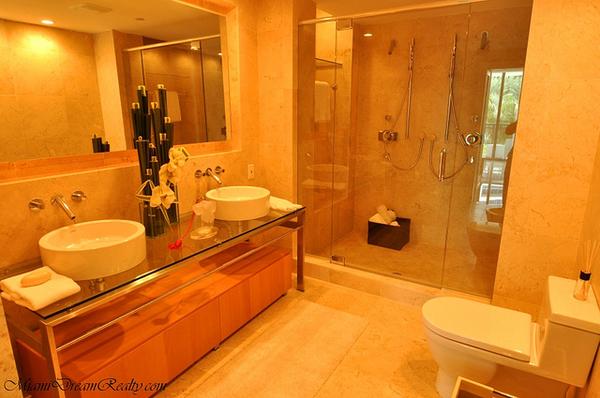} & \includegraphics[width=3cm,keepaspectratio]{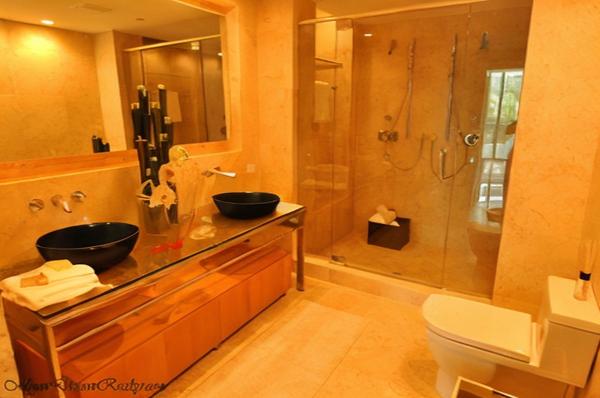}  \\ 
A couple of \textbf{white} bathroom sinks sitting next to a toilet. & A couple of \textbf{black} bathroom sinks sitting next to a toilet.  \\ \bottomrule
\end{tabular}
\caption{Color}
\label{tab:color}
\end{subfigure}%

\bigskip 

\begin{subfigure}[t]{0.48\textwidth}
\centering
\begin{tabular}{p{3cm}p{3cm}}
\toprule
\includegraphics[width=3cm,keepaspectratio]{assets/inpaintCOCO/img_11_image.jpg} & \includegraphics[width=3cm,keepaspectratio]{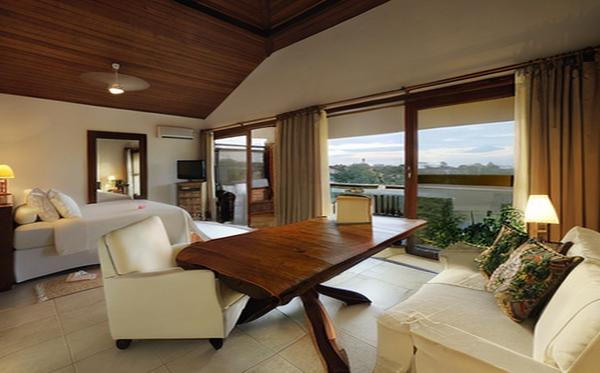}  \\ 
A living room with white furniture and a \textbf{small} wooden table. & A living room with white furniture and a \textbf{huge} wooden table.  \\ \bottomrule
\end{tabular}
\caption{Size}
\label{tab:size}
\end{subfigure}%

\caption{InpaintCOCO samples for all concepts.}
\label{tab:inpaintcoco}
\end{figure}

\end{document}